\documentclass[11pt]{article}

\usepackage[margin=1in]{geometry}
\usepackage[T1]{fontenc}
\usepackage[utf8]{inputenc}
\usepackage{lmodern}
\usepackage{microtype}
\usepackage{amsmath, amssymb, amsthm}
\usepackage{booktabs, tabularx, longtable, array}
\usepackage{enumitem}
\usepackage{xcolor}
\usepackage{ragged2e}
\usepackage[section]{placeins}
\usepackage{xurl}
\usepackage{hyperref}
\usepackage[numbers,sort&compress]{natbib}
\usepackage{graphicx}
\usepackage{tikz}
\usepackage{pgfplots}
\usetikzlibrary{arrows.meta, fit, positioning, shapes.geometric}
\pgfplotsset{compat=1.18}

\hypersetup{
  colorlinks=true,
  linkcolor=blue!50!black,
  citecolor=blue!50!black,
  urlcolor=blue!50!black,
  pdftitle={CAVA: Canonical Action Verification and Attestation},
  pdfauthor={Zexun Wang},
}

\setlist[itemize]{leftmargin=1.4em}
\setlist[enumerate]{leftmargin=1.6em}

\newcolumntype{L}[1]{>{\RaggedRight\arraybackslash}p{#1}}

\newtheorem{definition}{Definition}
\newtheorem{proposition}{Proposition}

\definecolor{cavaraw}{HTML}{B6B6B6}
\definecolor{cavatoken}{HTML}{F28C38}
\definecolor{cavaruntime}{HTML}{2E7DB6}
\definecolor{cavarisk}{HTML}{D9534F}

\pgfplotsset{
  cava chart/.style={
    width=0.94\linewidth,
    height=0.33\textheight,
    ymajorgrids=true,
    grid style={dashed, gray!25},
    axis line style={black!45},
    tick style={black!45},
    xlabel style={font=\small},
    ylabel style={font=\small},
    tick label style={font=\scriptsize},
    title style={font=\small},
    legend style={font=\scriptsize, draw=none, fill=none},
  },
}

\title{
  CAVA: Canonical Action Verification and Attestation\\
  for Runtime Governance of Agentic AI Systems
}
\author{Zexun Wang\thanks{Correspondence: \texttt{jw@nd.im}}\\Ond Holdings Inc.}
\date{June 2026}

\begin{document}

\maketitle

\begin{abstract}
Agentic AI systems increasingly act through heterogeneous runtimes: local coding hooks, SDK tools, browser automation, managed-agent traces, API gateways, and workflow engines. A single operational act such as publishing code, changing identity state, moving money, or exporting data may therefore be represented by many incompatible runtime records. This makes a basic governance question difficult to answer: what action was actually approved, what evidence binds the approval to execution, and can an independent verifier reproduce the same action identity later?

This paper presents \emph{Canonical Action Verification and Attestation} (CAVA), a runtime-semantics layer for converting heterogeneous agent activity into canonical runtime action objects. CAVA is positioned below Proof-Carrying Agent Actions (PCAA): PCAA defines the deployer-owned route-review-prove governance process, while CAVA defines the stable action object that process governs. The paper formalizes canonical action identity, semantic pattern detection, approval binding, receipt integrity, runtime-portable projection, and optional attestation substrates. We study a reference implementation through a 96-seed, 384-variant benchmark covering semantic equivalence, semantic separation, wrapper bypass, false-positive control, approval binding, receipt reproducibility, attestation tamper detection, runtime portability, semantic pattern detection, policy degradation, and Azure deployment drills. We further include a system-card appendix with ablations, red-team cases, comparative boundaries, deployment evidence, and residual-risk disclosures. In this representative corpus, CAVA preserves canonical action identity across rewritten runtime forms while raw-text and first-token baselines fail under wrappers, policy-addressable pattern detection, and approval drift. The contribution is a systems formulation of action-level canonicalization and policy-addressable semantic patterns as a necessary substrate for deployer-side AI governance.
\end{abstract}

\part{Main Research Manuscript}

\section{Introduction}

The operational risk of an agentic AI system is rarely realized at the point where a model emits prose. Risk becomes concrete when the runtime acts: a command is executed, a tool is invoked, a browser submits a form, a workflow mutates state, an identity boundary changes, a payment is triggered, a deployment leaves the workspace, or data crosses an organizational boundary. Enterprise governance therefore needs more than model-side policy, chat logs, or post-hoc traces. It needs a stable object that identifies the action being decided.

Today, that object is not stable. The same high-impact action may appear as a shell command in a local coding agent, an SDK method inside an agent framework, an MCP tool call, a browser automation event, a CI/CD API request, or a managed-agent session transition. Each representation may be useful for debugging, but none is guaranteed to be the governance object. If approval binds to raw text, an equivalent rewrite may bypass the approval. If audit binds to a runtime-native trace, a different runtime may make the same action incomparable. If policy binds to a first token, wrappers such as \texttt{env}, \texttt{sudo}, \texttt{bash -c}, aliases, SDK helpers, or tool indirection can change the surface without changing the consequence.

This paper studies CAVA as a runtime-semantics substrate for this problem. CAVA transforms raw runtime events into a \emph{canonical runtime action}: a versioned, hashable, and receipt-bearing action object. It is not a model-alignment method, a general observability product, or a replacement for enterprise policy. Instead, it addresses a narrower prerequisite:

\begin{quote}
Before a deployer can decide whether an agent action should proceed, the deployer needs a reproducible representation of what action is being decided.
\end{quote}

CAVA is designed to compose with PCAA \citep{pcaa2026}. PCAA defines the governance loop: route the action, review when needed, and prove closure. CAVA supplies the canonical action object that can travel through that loop. In short, PCAA answers who has authority and what proof must close the action; CAVA answers what exactly the authority decision refers to.

The paper makes five contributions:
\begin{enumerate}
  \item a formalization of canonical runtime action identity for heterogeneous agent systems;
  \item a CAVA protocol that binds policy outcomes, approvals, receipts, and attestations to canonical action fingerprints rather than raw text;
  \item a Semantic Pattern Layer inside CAVA that maps canonical actions and externality context into policy-addressable patterns rather than customer-specific rules;
  \item a threat model for semantic bypass, wrapper bypass, approval drift, evidence laundering, and parser capture;
  \item a reference open-core implementation boundary separating portable schema and receipt verification from managed parser packs and enterprise evidence operations;
  \item a reproducible benchmark harness with 96 representative seeds and 384 runtime variants that compares CAVA with raw-text and first-token baselines across action semantics, semantic pattern detection, policy degradation, cloud-action drills, ablations, and red-team cases.
\end{enumerate}

\section{Related Work}

\subsection{Agent evaluation and tool-use risk}

AgentBench frames agent evaluation as an interactive-environment problem rather than only a static-response problem \citep{agentbench2023}. SWE-bench similarly treats real software repositories as a practical testbed for evaluating autonomous coding behavior \citep{swebench2023}. ToolEmu focuses on risk discovery for tool-augmented language-model agents through emulated tool execution \citep{toolemu2023}. CAVA is aligned with the same premise that agent behavior must be evaluated at the action layer. Its target, however, is not task success or scenario discovery. Its target is governance-grade action identity: whether a runtime event can be canonicalized, approved, replayed, and attested under heterogeneous execution surfaces.

\subsection{Runtime governance, telemetry, and observability}

Runtime-control specifications and managed-agent systems increasingly expose sessions, tool calls, execution state, and enterprise controls. Microsoft Agent Control Specification emphasizes portable runtime governance contracts for agents \citep{microsoftacs2026}. OpenTelemetry GenAI semantic conventions standardize telemetry fields for LLM and agent systems \citep{otelgenai2026}. Langfuse and similar platforms expose tracing, prompt management, evaluations, and observability workflows \citep{langfuseobservability}. These systems make adjacent records legible. CAVA differs in the object it treats as primary: not a span, session, prompt, or trace, but a canonical action object suitable for approval binding and receipt verification.

\subsection{Proof, receipts, and attestations}

Proof-carrying code established the idea that an executable artifact can travel with a machine-checkable proof object \citep{necula1997pcc}. Modern software supply-chain systems such as in-toto and Sigstore provide signed provenance and attestation patterns for build and release workflows \citep{intoto2023,sigstore2024}. Verifiable Credentials provide portable claim structures for cross-party verification \citep{w3cvc2025}. Dapr Verifiable Execution contributes signed workflow history and execution attestation \citep{daprverifiable2026}. CAVA is complementary: it defines what runtime action those receipts or attestations should bind to.

\subsection{Governance frameworks}

Frontier-governance frameworks increasingly document risk identification, risk analysis, mitigations, incident response, model reporting, expert input, responsibility allocation, and change management \citep{openai2026fgf}. CAVA adopts the same operational seriousness but applies it to deployer-side runtime action semantics rather than frontier model release. The distinction matters because model providers can report capability and mitigation posture, but deployers still own production systems, customer data, enterprise identity boundaries, and business authority.

\section{Problem Formulation}

\subsection{Runtime heterogeneity}

Let $\mathcal{R}$ denote the set of runtime families through which an agent may act. Examples include shell hooks, SDK tools, API gateways, MCP servers, browser automation, managed-agent platforms, workflow engines, and Web3 signing lanes. A raw runtime event is denoted
\[
a \in \mathcal{A}_r \quad \text{for runtime } r \in \mathcal{R}.
\]
The deployer-facing question is not merely whether $a$ contains a suspicious string. The question is whether $a$ represents a business action that should be allowed, warned, approval-gated, blocked, or later audited.

\begin{definition}[Canonical runtime action]
For a raw runtime event $a$, a CAVA canonicalizer maps $a$ to a canonical runtime action
\[
C(a) = (v, r, e, o, k, S, \tau, u, m),
\]
where $v$ is schema version, $r$ is runtime family, $e$ is executable or tool identity, $o$ is normalized operation, $k$ is risk category, $S$ is the set of touched systems, $\tau$ is reversibility, $u$ is target or subject context, and $m$ is bounded adapter metadata.
\end{definition}

\begin{definition}[Canonical fingerprint]
Let $\mathrm{canon}(\cdot)$ denote deterministic serialization over selected canonical fields. The CAVA fingerprint is
\[
F(a) = H(\mathrm{canon}(C(a))),
\]
where $H$ is a collision-resistant hash function. Policy decisions, approval receipts, and audit evidence bind to $F(a)$ rather than to raw runtime text.
\end{definition}

\subsection{Desired properties}

CAVA is designed around six properties:
\begin{enumerate}
  \item \textbf{Semantic equivalence.} Equivalent expressions of the same action should converge to the same canonical identity.
  \item \textbf{Semantic separation.} Materially different actions should not collapse into the same governance object.
  \item \textbf{Wrapper robustness.} Shell wrappers, SDK helpers, tool aliases, and runtime indirection should not hide high-impact actions.
  \item \textbf{Approval binding.} Approval must bind to canonical action meaning, not to a display string.
  \item \textbf{Receipt reproducibility.} Independent verifiers should be able to recompute receipt hashes from the published object.
  \item \textbf{Runtime portability.} Different runtime families should project comparable action semantics without pretending that all runtimes expose equal enforcement depth.
\end{enumerate}

\subsection{Threat model}

The threat model includes both malicious and accidental failures:
\begin{itemize}
  \item \textbf{Equivalent syntax bypass}: the same action is rewritten so a raw-string policy no longer matches.
  \item \textbf{Wrapper bypass}: the action is placed behind \texttt{env}, \texttt{sudo}, \texttt{bash -c}, aliases, helper scripts, SDK calls, or tool indirection.
  \item \textbf{Approval drift}: an operator approves one surface description while a different semantic action executes.
  \item \textbf{Trace ambiguity}: observability records exist but do not prove what was authorized.
  \item \textbf{Evidence laundering}: a downstream system provides a sanitized after-the-fact record.
  \item \textbf{Parser capture}: a vendor-specific parser becomes the only governance authority.
\end{itemize}

CAVA does not assume that every runtime can provide inline interception. It assumes only that runtime events can be captured, normalized, scored for coverage depth, and attached to explicit receipts. When a runtime is observe-only, CAVA should disclose that limitation rather than overstate enforcement.

\section{Canonical Action Verification and Attestation}

\subsection{Protocol}

CAVA can be expressed as a six-stage protocol:
\begin{enumerate}
  \item \textbf{Capture}: collect the raw runtime event and execution context.
  \item \textbf{Normalize}: map the raw event into a canonical runtime action.
  \item \textbf{Interpret}: detect policy-addressable semantic patterns from canonical action, boundary, provenance, and data signals.
  \item \textbf{Fingerprint}: compute a deterministic hash over canonical action semantics.
  \item \textbf{Bind}: attach policy outcomes, approvals, denials, or escalations to the fingerprint. This is approval binding at the action-meaning layer.
  \item \textbf{Close}: attach outcome, evidence, exceptions, and side-effect summaries.
  \item \textbf{Attest}: optionally sign, credential, ledger-anchor, or externally verify receipt material.
\end{enumerate}

\begin{figure}[t]
\centering
\resizebox{\linewidth}{!}{%
\begin{tikzpicture}[
  node distance=0.62cm,
  box/.style={draw, rounded corners=3pt, align=center, minimum height=0.95cm, minimum width=2.15cm, text width=2.15cm, fill=blue!4, font=\small},
  receipt/.style={box, fill=green!5, minimum width=2.5cm, text width=2.5cm},
  arrow/.style={-{Latex[length=3mm]}, thick}
]
\node[box] (capture) {Capture\\raw event};
\node[box, right=of capture] (normalize) {Normalize\\action};
\node[box, right=of normalize] (interpret) {Interpret\\patterns};
\node[box, right=of interpret] (fingerprint) {Fingerprint\\semantics};
\node[box, right=of fingerprint] (bind) {Bind\\approval};
\node[box, right=of bind] (close) {Close\\evidence};

\draw[arrow] (capture) -- (normalize);
\draw[arrow] (normalize) -- (interpret);
\draw[arrow] (interpret) -- (fingerprint);
\draw[arrow] (fingerprint) -- (bind);
\draw[arrow] (bind) -- (close);

\node[receipt, below=0.9cm of fingerprint] (receipt) {Receipt hash\\and verifier};
\node[receipt, below=0.9cm of close] (attest) {Optional signer, VC,\\or ledger anchor};
\draw[arrow] (fingerprint) -- (receipt);
\draw[arrow] (close) -- (attest);
\end{tikzpicture}
}
\caption{CAVA protocol: raw runtime events are converted into canonical action objects, interpreted into policy-addressable semantic patterns, and bound to fingerprints before approval, closure, and optional attestation.}
\label{fig:cava-protocol}
\end{figure}
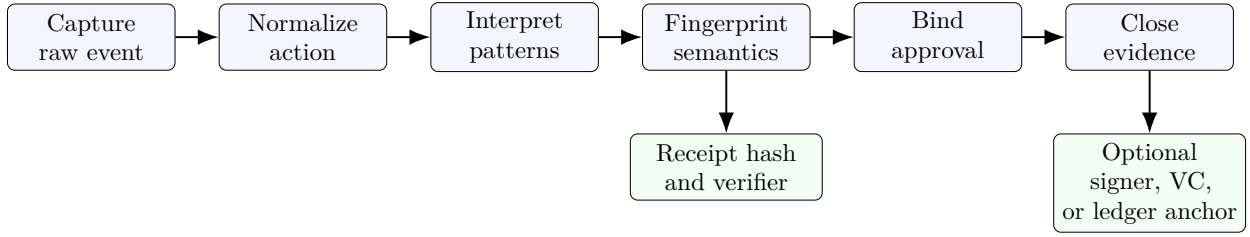

\subsection{Canonical action object}

Table~\ref{tab:canonical-action} gives the reference canonical object. The schema is intentionally small. It is not meant to encode every private workflow detail. It is meant to preserve the fields required for governance-grade action identity.

\begin{table}[t]
\centering
\small
\caption{Reference CAVA canonical action fields.}
\label{tab:canonical-action}
\begin{tabularx}{\linewidth}{@{}>{\raggedright\arraybackslash}p{0.22\linewidth}>{\raggedright\arraybackslash}p{0.24\linewidth}X@{}}
\toprule
Field & Example & Role in governance \\
\midrule
\texttt{schema\_version} & \texttt{osuite.cava...v1} & Versioned interoperability boundary \\
\texttt{runtime} & \texttt{bash}, \texttt{mcp}, \texttt{browser} & Runtime family and coverage context \\
\texttt{adapter} & \texttt{codex-hook}, \texttt{hosted-agent} & Capture mechanism and trust depth \\
\texttt{executable} & \texttt{git}, \texttt{stripe}, \texttt{kubectl} & Tool or executable identity \\
\texttt{operation} & \texttt{push}, \texttt{payment\_trigger} & Normalized operation \\
\texttt{category} & \texttt{deployment}, \texttt{payment} & Risk taxonomy input \\
\texttt{systems\_touched} & \texttt{github}, \texttt{postgres} & Blast-radius input \\
\texttt{reversible} & \texttt{false} & Approval and rollback input \\
\texttt{target}/\texttt{subject} & branch, account, tenant, wallet & Affected object or boundary \\
\bottomrule
\end{tabularx}
\end{table}

\begin{proposition}[Raw text is insufficient for approval binding]
If a governance decision binds only to a raw string representation $s(a)$, then any alternative representation $s'(a)$ with the same operational effect but different text may evade a policy keyed on $s(a)$ unless the policy independently reconstructs action semantics.
\end{proposition}

\noindent The proposition is not a cryptographic theorem; it is a systems observation. Raw text is a display form. CAVA makes the reconstructed action semantics the governance object.

\subsection{Relationship to PCAA}

PCAA treats a governed action as a certificate-bearing object with checkpoints for admissibility, action open, assumption capture, approval, and outcome closure \citep{pcaa2026}. CAVA is the lower layer that makes the object stable enough to govern. A PCAA certificate can contain route decisions and proof bundles, but those decisions need a canonical action identity to avoid ambiguity. Conversely, a CAVA fingerprint alone does not decide authority. It becomes operationally meaningful when a PCAA or equivalent deployer-owned governance loop routes, reviews, and closes it.

\subsection{Semantic Pattern Layer}

Canonical action identity is necessary but not sufficient. Enterprise policy rarely says only ``block this executable.'' It says things like: do not create public links for sensitive material, do not weaken endpoint controls, do not let an agent exercise authority outside the user's delegated scope, and do not let workflow sinks move data into public channels. These are not vendor-specific commands; they are reusable agentic risk patterns. CAVA therefore includes a \emph{Semantic Pattern Layer} that sits after canonicalization but before policy routing.

The pattern layer is not a new product and not a second governance authority. It is an internal CAVA interpretation pass. CAVA owns structure; semantic patterns own risk meaning; policy profiles own enterprise posture; PCAA owns final authority and proof. Formally, a pattern detector maps
\[
  P(C(a), B(a), D(a)) \rightarrow \{p_1,\ldots,p_n\},
\]
where $C(a)$ is the canonical action, $B(a)$ is boundary context such as destination visibility and account provenance, and $D(a)$ is data context such as sensitivity and minimization posture. Each pattern is a versioned evidence object with identifier, label, severity, confidence, profile axis, minimum decision, and evidence tuples.

\begin{table}[t]
\centering
\small
\caption{Reference CAVA semantic patterns.}
\label{tab:cava-patterns}
\begin{tabularx}{\linewidth}{@{}L{0.25\linewidth}L{0.25\linewidth}X@{}}
\toprule
Pattern & Policy axis & Meaning \\
\midrule
Hidden externality & External sharing & A normal-looking task creates an external side effect such as an upload, share, vendor handoff, or public URL. \\
Public persistent egress & Public links & Data moves to a public or link-accessible destination that may persist outside the workspace. \\
Security-control weakening & Production changes & The action disables or weakens endpoint, firewall, audit, EDR, runtime, or governance controls. \\
Credential exposure & External sharing & Secret, token, credential, or key material may leave a trusted boundary. \\
Delegated authority mismatch & Identity admin & The agent exercises identity, OAuth, service-account, or delegated authority beyond the visible task boundary. \\
Workflow sink risk & External sharing & A graph or workflow sends data to a public, external, or weakly governed sink. \\
Prompt or rule tampering & Production changes & The action changes prompts, rules, managed settings, policy files, or hook behavior that shape future actions. \\
\bottomrule
\end{tabularx}
\end{table}

This distinction prevents the system from becoming a pile of customer-specific exceptions. For example, a public file handoff should not be encoded as ``block one named file host.'' It should be encoded as public persistent egress with evidence: destination type, visibility, persistence, account provenance, and payload sensitivity. A policy profile can then route the same pattern differently for a startup sandbox, an incident-response workspace, or a regulated enterprise deployment.

The pattern layer also aligns CAVA with emerging agentic-risk taxonomies without reducing CAVA to a checklist. OWASP's agentic AI work emphasizes risks such as tool misuse, excessive agency, insecure execution, and memory or context abuse, while NIST's generative AI profile frames risk management as a context-sensitive governance activity rather than a single universal rule set \citep{owaspagentic2026,nistai6001}. CAVA's role is narrower: convert runtime behavior into evidence-bearing patterns that policy profiles can address.

\subsection{Bounded Action Firewall and runtime exposure graph}

Recent implementation work separates two higher-level primitives from CAVA while keeping them dependent on CAVA's canonical action object. The first is a \emph{Bounded Action Firewall}: a runtime gate that returns allow, ask, block, or observe only after binding the decision to a canonical fingerprint, policy version, actor identity, runtime session, destination scope, proof receipt, and time window. In this model, a human approval is not a broad permission attached to a chat session. It is an action gate lease for a specific semantic action. If the target, policy, actor, runtime session, or proof digest changes, the lease must expire or require re-approval.

The second is an \emph{Agent Runtime Exposure Graph}: a graph projection that connects agents, runtime adapters, sessions, canonical actions, policy profiles, final-authority decisions, destinations, data boundaries, proof receipts, and incident events. CAVA supplies the action nodes and fingerprints; PCAA supplies the authority and closure edges; the bounded firewall supplies gate outcomes; the exposure graph supplies reachability, blast-radius, incident reconstruction, and framework-mapping queries.

These primitives are deliberately not defined as new product lines. They are productized consequences of the CAVA/PCAA stack. CAVA remains the semantic action layer. The bounded firewall is the enforcement layer that makes approval replay-resistant. The exposure graph is the security-operations layer that makes blast radius and evidence legible to CISOs. This separation avoids overloading CAVA with every downstream product concern while preserving the scientific claim that canonical action identity is the object on which runtime governance depends.

\begin{table}[t]
\centering
\small
\caption{How CAVA composes with BAF and AREG inside OSuite.}
\label{tab:cava-baf-areg}
\begin{tabularx}{\linewidth}{@{}L{0.22\linewidth}L{0.28\linewidth}X@{}}
\toprule
Layer & Primary object & Responsibility \\
\midrule
CAVA & Canonical action fingerprint & Reconstruct action meaning from heterogeneous runtime events. \\
Policy profile & Enterprise posture & Decide how semantic patterns should route in this workspace. \\
Decision Score & Risk feature vector & Rank impact, exposure, control weakness, and evidence confidence. \\
PCAA & Final authority and proof loop & Decide who may approve and what proof closes the action. \\
Bounded Action Firewall & Action gate lease & Enforce allow, ask, block, or observe with replay-resistant approval binding. \\
AREG & Runtime exposure graph & Map blast radius, boundary drift, framework coverage, and incident evidence. \\
\bottomrule
\end{tabularx}
\end{table}

\subsection{Attestation substrates}

CAVA receipts can remain local hash receipts, or they can be extended with stronger substrates:
\begin{itemize}
  \item local deterministic SHA-256 receipt hashes;
  \item workspace, runtime, or operator signatures;
  \item verifiable credentials for external procurement and auditor review;
  \item in-toto or Sigstore-style supply-chain attestations;
  \item permissioned or public ledger anchoring of receipt digests;
  \item smart-account policies for wallet, relayer, or settlement lanes.
\end{itemize}

The design intentionally treats blockchain and Web3 components as optional attestation substrates, not as the definition of CAVA. Public ledgers may provide timestamping, non-repudiation, or settlement finality, but confidential enterprise action contents should not be forced on-chain.

\section{Reference Implementation}

The current reference implementation has two layers. The OSuite implementation contains production-facing CAVA parsing, PCAA policy integration, approval workflow binding, evidence graph projection, and enterprise assurance surfaces. The open CAVA package exposes a smaller skeleton: schema constants, deterministic hashing, receipt creation and verification, profile normalization, and runtime-adapter contracts.

This split is deliberate. The open package should be sufficient for researchers and developers to reproduce the core semantics and test adapters. It should not replicate OSuite's managed parser packs, enterprise policy routing, buyer-ready evidence graph, signer orchestration, private KMS/HSM integration, or managed connectors. This open-core boundary reduces lock-in while preserving a commercial reason to buy the managed product.

\begin{table}[t]
\centering
\small
\caption{Open-core boundary in the reference implementation.}
\begin{tabularx}{\linewidth}{@{}>{\raggedright\arraybackslash}p{0.28\linewidth}X@{}}
\toprule
Layer & Contents \\
\midrule
Open CAVA skeleton & Schema, deterministic fingerprints, receipt build/verify helpers, adapter contracts, local examples \\
OSuite managed layer & Production parser packs, PCAA policy routing, approval binding workflows, evidence graph, replay, exports, enterprise signers, KMS/HSM, optional VC or ledger anchoring \\
\bottomrule
\end{tabularx}
\end{table}

\section{Evaluation}

\subsection{Benchmark design}

We evaluate CAVA with a disclosure-safe benchmark harness included with the manuscript artifacts. The public harness contains 96 representative seed scenarios and expands them into 384 runtime variants across four runtime families: shell hooks, MCP-style tools, browser automation, and managed-agent traces. Production parser packs, enterprise policy thresholds, customer connector rules, and the managed OSuite evidence graph are withheld.

The benchmark tests nine properties:
\begin{enumerate}
  \item semantic equivalence across rewritten action forms;
  \item semantic separation across materially different operations;
  \item wrapper-bypass catch rate;
  \item false-positive control for benign text containing high-impact strings;
  \item approval binding correctness;
  \item receipt reproducibility;
  \item attestation tamper detection;
  \item runtime portability convergence;
  \item semantic pattern detection and policy-profile routing.
\end{enumerate}

The baselines are intentionally simple but operationally common:
\begin{itemize}
  \item \textbf{Raw-text policy}: binds to the literal runtime string.
  \item \textbf{First-token rules}: classifies actions using the first executable and neighboring token.
  \item \textbf{CAVA runtime}: normalizes wrappers, aliases, categories, and receipt material before decision binding.
\end{itemize}

\subsection{Results}

Table~\ref{tab:cava-results} reports aggregate results from the current harness. The numbers should be read as regression evidence for the representative corpus, not as a universal claim over all future enterprise runtimes. The public benchmark deliberately separates two layers: scored executable checks for canonicalization and receipt integrity, and structured system-card material for policy degradation, cloud-action drills, ablations, and red-team case analysis.

\begin{table}[t]
\centering
\small
\caption{Aggregate benchmark results on the 96-seed, 384-variant CAVA corpus.}
\label{tab:cava-results}
\begin{tabularx}{\linewidth}{@{}>{\raggedright\arraybackslash}p{0.35\linewidth}ccc@{}}
\toprule
Metric & Raw text & First-token & CAVA \\
\midrule
Semantic equivalence recall & 0.000 & 0.000 & \textbf{1.000} \\
Semantic separation precision & 1.000 & 0.750 & \textbf{1.000} \\
Wrapper-bypass catch rate & 0.000 & 0.000 & \textbf{1.000} \\
False-positive control & 0.500 & 0.500 & \textbf{1.000} \\
Approval binding correctness & 0.000 & 0.000 & \textbf{1.000} \\
Receipt reproducibility & 0.000 & 0.000 & \textbf{1.000} \\
Attestation tamper detection & 0.000 & 0.000 & \textbf{1.000} \\
Runtime portability convergence & 0.000 & 0.000 & \textbf{1.000} \\
Semantic pattern detection & 0.000 & 0.333 & \textbf{1.000} \\
\bottomrule
\end{tabularx}
\end{table}

\begin{figure}[t]
\centering
\begin{tikzpicture}
\begin{axis}[
  cava chart,
  title={CAVA vs. Raw-Text and First-Token Baselines},
  ylabel={Score},
  ymin=0,
  ymax=1.05,
  ybar,
  bar width=9pt,
  symbolic x coords={Equivalence,Wrapper,False Positive,Patterns,Approval,Attestation},
  xtick=data,
  enlarge x limits=0.12,
  legend columns=3,
  legend style={at={(0.5,1.13)}, anchor=south},
]
\addplot[fill=cavaraw, draw=cavaraw] coordinates {
  (Equivalence,0)
  (Wrapper,0)
  (False Positive,0.5)
  (Patterns,0)
  (Approval,0)
  (Attestation,0)
};
\addplot[fill=cavatoken, draw=cavatoken] coordinates {
  (Equivalence,0)
  (Wrapper,0)
  (False Positive,0.5)
  (Patterns,0.333)
  (Approval,0)
  (Attestation,0)
};
\addplot[fill=cavaruntime, draw=cavaruntime] coordinates {
  (Equivalence,1)
  (Wrapper,1)
  (False Positive,1)
  (Patterns,1)
  (Approval,1)
  (Attestation,1)
};
\legend{Raw text, First-token, CAVA}
\end{axis}
\end{tikzpicture}
\caption{Representative benchmark slices. CAVA's advantage comes from binding decisions to canonical action semantics rather than display text or first-token heuristics.}
\label{fig:cava-benchmark}
\end{figure}
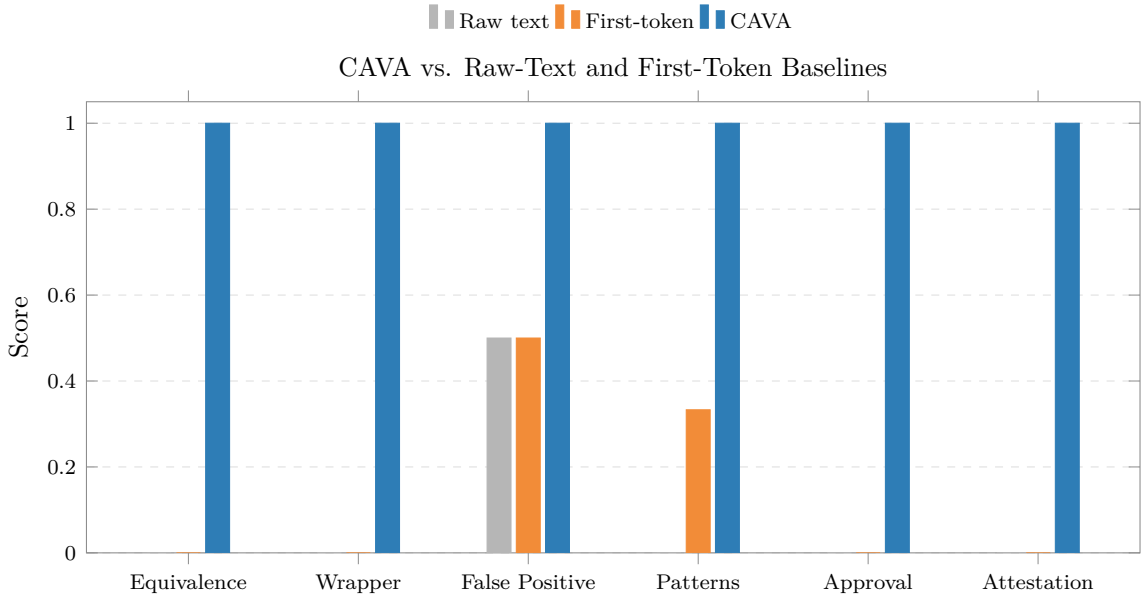

\section{Expanded Evaluation Matrix}

The benchmark is intentionally not a single score. A single aggregate number would hide the failure modes that matter to buyers and reviewers: whether wrappers are handled, whether benign text is over-blocked, whether approvals bind to the executed action, whether receipts are reproducible, whether runtime families converge, and whether governance itself can be weakened. Table~\ref{tab:evaluation-matrix} reports the evaluation matrix used by the artifact package.

\begin{longtable}{@{}L{0.20\linewidth}L{0.33\linewidth}L{0.17\linewidth}L{0.18\linewidth}@{}}
\caption{Expanded CAVA evaluation matrix. The support column includes scored checks and structured system-card cases in the public artifact.}
\label{tab:evaluation-matrix}\\
\toprule
Suite & What is challenged & Public support & Primary metric \\
\midrule
\endfirsthead
\toprule
Suite & What is challenged & Public support & Primary metric \\
\midrule
\endhead
Semantic equivalence & Same operational action rewritten through flags, environment variables, command wrappers, or helper forms. & 12 scored variants & Equivalence recall \\
Semantic separation & Materially different actions that share words, tools, or nearby context. & 8 pairwise checks & Separation precision \\
Wrapper bypass & High-impact action hidden behind \texttt{env}, \texttt{sudo}, \texttt{command}, \texttt{bash -c}, or nested shells. & 14 checks including ablations & Bypass catch rate \\
Benign contamination & Read-only commands containing high-impact strings such as \texttt{git push}, \texttt{kubectl delete}, or payment events. & 7 checks including ablations & False-positive control \\
Approval drift & Receipt replay or approval reuse after canonical fields change. & 8 checks including ablations & Approval-binding correctness \\
Receipt tampering & Modified policy outcomes, altered receipt fields, or order-dependent serialization. & 8 checks & Hash reproducibility and tamper detection \\
Runtime portability & Shell, MCP, browser, and managed-agent records describing comparable operational actions. & 16 runtime projections & Portability convergence \\
Semantic pattern detection & Canonical actions plus boundary context mapped to reusable patterns such as public egress, security-control weakening, delegated authority mismatch, and workflow sink risk. & 6 scored cases & Pattern detection and profile routing \\
Policy degradation & Signature disabling, permissive policy changes, trust-boundary broadening, and evidence retention weakening. & 8 structured cases & Degradation coverage \\
Azure deployment drill & Azure CLI actions across what-if, deployment, role assignment, Key Vault, SQL firewall, and ACR read paths. & 6 structured cases & Cloud-action projection coverage \\
Red-team casebook & Attacker-style narratives linking bypass method, baseline failure, expected CAVA behavior, and residual risk. & 24 cases & Qualitative coverage \\
\bottomrule
\end{longtable}

This matrix is the paper's answer to a common objection against governance components: that they look good only when evaluated on the exact examples used to explain them. The CAVA harness separates \emph{seed scenario}, \emph{runtime variant}, \emph{suite}, and \emph{evidence type}. A seed may be scored directly, used as an ablation probe, or carried into the red-team casebook. The public corpus is still small compared with enterprise reality, but it is structured so that new runtime adapters and customer-approved traces can be added without changing the metric vocabulary.

\section{Ablation Study}

Table~\ref{tab:ablation-study} reports the ablation study used in the system-card artifact. These ablations are not meant to claim that the exact implementation is optimal. They identify the components that are necessary for the CAVA claim to be true. If any of these layers is removed, CAVA collapses into either ordinary logging, brittle guardrails, or a product-specific approval workflow.

\begin{longtable}{@{}L{0.22\linewidth}L{0.27\linewidth}L{0.32\linewidth}L{0.09\linewidth}@{}}
\caption{CAVA ablation study. Retained score is the expected retained capability for the affected suite under the public harness.}
\label{tab:ablation-study}\\
\toprule
Ablation & Affected suites & Expected failure & Retained score \\
\midrule
\endfirsthead
\toprule
Ablation & Affected suites & Expected failure & Retained score \\
\midrule
\endhead
Remove wrapper parsing & Wrapper bypass, semantic equivalence & \texttt{env}, \texttt{sudo}, \texttt{bash -c}, and nested shells remain surface text; high-impact action can hide behind the wrapper. & 0.00 \\
Remove profile aliases & Payment, Web3, customer private tools & Domain-specific operations such as Stripe triggers or wallet sends collapse to unknown or low-confidence actions. & 0.50 \\
Remove canonical fingerprint & Approval drift, runtime portability & Approval binds to display text or runtime-native IDs rather than action meaning. & 0.00 \\
Remove receipt verifier & Receipt tampering, external assurance & Changed decisions and receipt payloads become log entries instead of verifier failures. & 0.00 \\
Remove PCAA binding & Policy degradation, approval drift, outcome closure & Canonical fingerprints exist, but no deployer-owned route-review-prove loop decides authority or closure. & 0.25 \\
\bottomrule
\end{longtable}

The harshest ablations are canonical fingerprint removal and receipt-verifier removal. Without a canonical fingerprint, there is no stable object for approval. Without verifier logic, the receipt becomes decorative evidence. This is why CAVA should not be marketed as merely an observability schema. Observability can tell a team that something happened. CAVA must help a team prove what was authorized, what actually happened, and whether the evidence survived replay.

\section{Interpretation}

The benchmark demonstrates three points. First, raw-text governance has no stable notion of semantic equivalence: \texttt{git push origin main}, \texttt{git -c push.default=simple push origin main}, \texttt{env ... git push origin main}, and \texttt{bash -c "git push origin main"} are different strings but the same governed action. Second, first-token rules do not survive wrappers and are prone to false positives when high-impact strings appear in search, documentation, or echo commands. Third, canonical receipts allow approval and attestation checks that raw strings do not provide.

The most important result is not that CAVA scores 1.0 on a controlled corpus. The important result is that the benchmark exposes the failure modes that a serious market-facing component must survive. Future versions should expand the corpus with third-party runtime traces, customer-approved anonymized examples, and adversarial parser challenges.

\section{Discussion}

\subsection{Why CAVA should not be only a product feature}

CAVA is more credible as a portable component than as a hidden OSuite feature. A public schema and verifier allow the ecosystem to inspect the core claim, reproduce hashes, and write adapters. OSuite can still monetize the hard parts: maintained parser coverage, enterprise approvals, evidence graph operations, assurance exports, support, signer orchestration, and managed integrations. This is the right open-core boundary for both adoption and revenue.

\subsection{Where Web3 belongs}

Web3 should be used carefully. If CAVA is defined as blockchain, it becomes narrower and less enterprise-friendly. If CAVA treats ledgers, verifiable credentials, smart accounts, and decentralized identity as optional attestation lanes, it gains stronger integrity options without forcing every deployer into a settlement substrate. The principle is simple: canonical action semantics first; attestation substrate second.

\subsection{Market testability}

For CAVA to survive market scrutiny, the benchmark must be uncomfortable. It should include not only clean examples but wrappers, nested shells, alias abuse, SDK indirection, browser-side mutations, MCP tool calls, hosted traces, benign commands containing dangerous strings, and tampered receipts. This paper introduces that test direction, but the component should be judged by ongoing benchmark expansion rather than by a static whitepaper.

\section{Threats to Validity}

\textbf{Representative corpus.} The current benchmark is a representative public corpus, not a complete distribution of enterprise actions. It is useful for regression and public explanation, but future evidence should include broader traces.

\textbf{Parser coverage.} Canonicalization is only as strong as adapter coverage. Private enterprise systems may need custom parser packs, and parser failures can produce either missed escalations or unnecessary review.

\textbf{Disclosure boundary.} Production parser packs and customer connector rules are withheld. This protects commercial and security-sensitive implementation details, but it limits full third-party reproduction of OSuite's managed behavior.

\textbf{Attestation scope.} Receipt hashes and signatures prove integrity and binding, not business wisdom. A correctly attested action can still be a bad decision.

\textbf{Runtime depth.} Some runtimes allow inline blocking. Others only provide observer-mode traces. CAVA must disclose that difference rather than claim uniform enforcement.

\section{Conclusion}

This paper argues that agent governance needs a stable action-semantics layer. Without it, approvals bind to strings, traces bind to runtime-specific records, and evidence becomes difficult to replay across changing execution surfaces. CAVA provides a canonical runtime action object, deterministic fingerprint, receipt path, and optional attestation architecture that can sit below PCAA or a similar deployer-owned governance loop.

The current reference implementation and benchmark are early but concrete. They show how CAVA can collapse equivalent runtime forms, separate materially different actions, resist wrappers, bind approvals to fingerprints, reproduce receipts, detect tampering, and project comparable semantics across runtime families. The next research step is to expand the public corpus, invite adversarial parser challenges, and evaluate CAVA against real third-party agent traces. The product step is equally clear: keep the schema open, keep the managed evidence layer worth paying for, and make runtime action governance measurable rather than rhetorical.

\appendix

\part{CAVA System Card and Technical Appendix}

\section{System Scope and Release Posture}

This appendix uses a system-card style to document CAVA as an evaluable governance component rather than as a finished safety guarantee. The system under study is the CAVA runtime-semantics layer and the surrounding OSuite reference implementation. The public open-core package exposes schema, deterministic hashing, receipts, profile normalization, and runtime adapter contracts. The managed OSuite layer adds production parser packs, PCAA policy routing, approval workflows, evidence graph construction, replay, exports, enterprise signers, private key-management integrations, verifiable credential issuance, optional ledger anchoring, and managed connectors.

The release posture is intentionally split. The open package is suitable for experimentation, adapter prototypes, reproducible benchmark execution, and third-party inspection of core hash and receipt semantics. It is not intended to replace a managed runtime governance system. The managed layer is intended for production deployments where parser maintenance, evidence retention, operator workflow, tenant security, and buyer-facing assurance matter.

\section{Observed Runtime Governance Challenges}

The reference implementation and benchmark are organized around observed failure modes in agent runtimes. These are not model-output categories such as toxicity or hallucination. They are operational governance failures that occur when an action crosses a runtime boundary.

\begin{longtable}{@{}L{0.23\linewidth}L{0.37\linewidth}L{0.28\linewidth}@{}}
\toprule
Challenge & Failure pattern & CAVA evaluation slice \\
\midrule
Equivalent syntax & The same operation is rewritten through flags, environment variables, shell wrappers, or helper commands. & Semantic equivalence recall \\
Semantic separation & A benign action contains a high-impact string or shares an executable with a high-impact action. & Semantic separation precision and false-positive control \\
Wrapper bypass & The high-impact operation hides behind \texttt{env}, \texttt{sudo}, \texttt{command}, \texttt{bash -c}, SDK indirection, or tool aliasing. & Wrapper-bypass catch rate \\
Approval drift & The operator approves display text while execution binds to a different action. & Approval binding correctness \\
Receipt drift & A receipt is recomputed under a different field order, profile, or decision payload. & Receipt reproducibility \\
Attestation tampering & A receipt or policy outcome is modified after approval. & Attestation tamper detection \\
Runtime fragmentation & Shell hooks, MCP tools, browser actions, and managed traces describe the same action differently. & Runtime portability convergence \\
\bottomrule
\end{longtable}

\section{Risk Taxonomy}

CAVA classifies runtime action risk by operational consequence rather than by model intent. The taxonomy is intended to remain small enough to be portable and large enough to support enterprise routing.

\begin{longtable}{@{}L{0.18\linewidth}L{0.33\linewidth}L{0.38\linewidth}@{}}
\toprule
Risk class & Examples & Governance implication \\
\midrule
Observation & Search, list, inspect, read-only query & Usually allow or log, unless sensitive data boundary is crossed. \\
Build and preparation & Build, dry run, plan, simulation & Often allow or simulate-first; useful for reducing unnecessary review. \\
Deployment & Push, release, publish, migrate & Requires canonical identity, approval binding, and outcome closure. \\
Infrastructure change & Terraform, Kubernetes, Helm, Pulumi, firewall change & Requires high-integrity receipts and rollback evidence. \\
Database mutation & Insert, update, delete, truncate, migration & Requires target clarity, reversibility disclosure, and approval binding. \\
Identity and authority & Key rotation, permission grant, user creation & Requires authority lineage and dual-control in high-impact cases. \\
Payment and obligation & Charge, refund, invoice, transfer & Requires receipt integrity and financial-system boundary disclosure. \\
Data boundary & Export, upload, email, public share & Requires destination visibility and account provenance. \\
Web3 settlement & Sign, send, bridge, swap, sponsor, relay & Requires cryptographic intent binding and optional ledger anchoring. \\
\bottomrule
\end{longtable}

\section{Capability Taxonomy}

The system-card posture for CAVA is capability-based rather than vendor-based. A deployment should not claim that ``CAVA is enabled'' as a binary statement. It should disclose what classes of runtime action can be captured, normalized, verified, and enforced. Table~\ref{tab:capability-taxonomy} defines the capability taxonomy used by the current paper and artifact package.

\begin{longtable}{@{}L{0.18\linewidth}L{0.27\linewidth}L{0.28\linewidth}L{0.18\linewidth}@{}}
\caption{CAVA capability taxonomy for deployment disclosure.}
\label{tab:capability-taxonomy}\\
\toprule
Capability & Minimum evidence & Failure if absent & Typical release posture \\
\midrule
\endfirsthead
\toprule
Capability & Minimum evidence & Failure if absent & Typical release posture \\
\midrule
\endhead
Capture & Raw event, runtime family, adapter ID, timestamp, actor or session handle. & The action may occur outside the evidence boundary. & Observe \\
Normalization & Canonical action fields and schema version. & Similar actions remain incomparable across runtime surfaces. & Observe or warn \\
Semantic separation & Evidence that benign lookalikes do not collapse into high-impact actions. & False positives erode operator trust and produce alert fatigue. & Warn \\
Wrapper resilience & Evidence that common wrappers are stripped or decomposed. & High-impact actions hide behind shell or SDK indirection. & Approval-gate \\
Approval binding & Receipt connecting decision, policy, and canonical fingerprint. & The operator may approve one thing while another executes. & Approval-gate \\
Receipt reproducibility & Deterministic serialization and verifier output. & Evidence cannot be independently replayed. & Approval-gate \\
Outcome closure & Execution result, exception, or partial-coverage marker. & Governance ends at approval and loses operational truth. & Approval-gate \\
Coverage disclosure & Enforcement depth, adapter confidence, and unknown-field posture. & Observe-only traces are confused with blocking controls. & Warn or approval-gate \\
Policy degradation detection & Canonical treatment of policy, signature, retention, and trust changes. & Governance controls can be weakened under the language of configuration. & Dual control \\
External attestation & Signature, verifiable credential, supply-chain attestation, or optional ledger anchor. & Buyer-facing assurance depends on local logs only. & External assurance \\
\bottomrule
\end{longtable}

This taxonomy is important for sales as much as for science. Buyers do not merely ask whether an AI governance product has policies. They ask whether the product can explain what it actually sees, what it can block, what it can only record, and what evidence survives after a disputed event. CAVA therefore treats \emph{capability claims} as auditable objects. A SaaS deployment may expose managed coverage and standard policies. A self-hosted enterprise deployment may add private parser packs, internal connectors, custom signer policies, and higher-friction approval routes. The open-core package can validate receipts and fingerprints, but it should not imply that a user has reproduced OSuite's managed parser, policy, evidence, or assurance layer.

\section{Safeguard Design}

CAVA safeguards are layered so that deployments can increase assurance without redefining the base object.

\begin{enumerate}
  \item \textbf{Normalization safeguards}: adapters strip common wrappers, resolve aliases, and project runtime-specific fields into stable canonical fields.
  \item \textbf{Binding safeguards}: policy outcomes and approvals attach to canonical fingerprints rather than raw strings.
  \item \textbf{Receipt safeguards}: receipt hashes are deterministic and can be recomputed by independent verifiers.
  \item \textbf{Coverage safeguards}: runtime family, adapter mode, and enforcement depth remain explicit so observer-only coverage is not presented as inline blocking.
  \item \textbf{Attestation safeguards}: selected receipt digests can be signed, credentialed, or ledger-anchored without publishing confidential action contents.
  \item \textbf{Commercial-boundary safeguards}: the open package exposes verification primitives, while managed OSuite holds production parser maintenance, evidence operations, and enterprise connector depth.
\end{enumerate}

\section{Red-Team and Adversarial Evaluation Plan}

The benchmark should evolve from representative examples into an adversarial corpus. The red-team plan is divided into six tracks:

\begin{enumerate}
  \item \textbf{Syntax rewriting}: equivalent operations are rewritten through flags, environment variables, shells, aliases, and scripts.
  \item \textbf{Benign contamination}: low-risk commands intentionally contain dangerous substrings such as \texttt{git push}, \texttt{kubectl delete}, or \texttt{DROP TABLE}.
  \item \textbf{Runtime projection}: the same action is represented as shell, MCP, browser, managed-agent, and API-gateway events.
  \item \textbf{Approval mismatch}: approval receipts are replayed against altered fingerprints or changed policy payloads.
  \item \textbf{Evidence tampering}: receipt fields are changed after closure to test verifier detection.
  \item \textbf{Parser miss discovery}: unknown tools, nested commands, customer-specific scripts, and private runtime events are used to identify coverage gaps.
\end{enumerate}

This plan is intentionally uncomfortable. A useful CAVA benchmark should make bypass attempts cheap to write, deterministic to reproduce, and hard for the canonicalizer to pass accidentally.

\section{Red-Team Casebook}

Table~\ref{tab:red-team-casebook} is a disclosure-safe red-team casebook extracted from the benchmark harness. The cases are intentionally concrete because vague safety prose is cheap; runtime governance becomes meaningful only when the attempted bypass can be reproduced. Each row records an attack pattern, a representative event, the baseline failure, the expected CAVA behavior, and residual risk. The casebook is not exhaustive. It is a forcing function for reviewers, customers, and future contributors to add uncomfortable examples rather than only success stories.

\footnotesize
\begin{longtable}{@{}L{0.07\linewidth}L{0.18\linewidth}L{0.23\linewidth}L{0.22\linewidth}L{0.20\linewidth}@{}}
\caption{Disclosure-safe CAVA red-team casebook.}
\label{tab:red-team-casebook}\\
\toprule
ID & Attack & Representative event & Baseline failure & CAVA expected behavior \\
\midrule
\endfirsthead
\toprule
ID & Attack & Representative event & Baseline failure & CAVA expected behavior \\
\midrule
\endhead
RT-001 & Nested shell deploy & \texttt{bash -lc npm-deploy} & First-token rules see \texttt{bash}, not deploy. & Unwrap nested shell and classify deployment. \\
RT-002 & Benign documentation search & \texttt{rg "kubectl delete namespace prod"} & Substring policy escalates read-only search. & Classify observation and suppress false positive. \\
RT-003 & Approval replay against changed branch & \texttt{git push origin release-candidate} & Display resembles approved push. & Reject receipt if fingerprint differs. \\
RT-004 & SDK payment indirection & \texttt{mcp.stripe.refund(...)} & No shell token exists to match. & Project MCP call into payment category. \\
RT-005 & Browser release button & \texttt{browser publish-release} & Trace only says click. & Bind UI action to deployment semantics. \\
RT-006 & Managed trace field drift & \texttt{agent github.push trace} & Runtime-native trace misses shell policy. & Project managed trace into deployment class. \\
RT-007 & Receipt decision tamper & \texttt{receipt.decision = "allow"} & Log-only systems may not recompute integrity. & Detect receipt-hash mismatch. \\
RT-008 & Policy loosening after import & \texttt{osuite policy update --mode allow-all} & Configuration looks administrative. & Classify as policy degradation. \\
RT-009 & Signature mode disable & \texttt{osuite signature off} & Verifier disappears without side-effect label. & Mark trust degradation and refuse proof claims. \\
RT-010 & Azure role escalation & \texttt{az role create Contributor} & Generic CLI command hides authority change. & Require identity-authority review. \\
RT-011 & Azure destructive Key Vault operation & \texttt{az keyvault purge --name prod-vault} & Irreversibility may be unknown. & Mark destructive and non-reversible. \\
RT-012 & Database mutation through MCP query & \texttt{mcp.postgres delete} & Generic query hides mutation. & Parse SQL operation and classify mutation. \\
RT-013 & Wallet settlement action & \texttt{cast send 0xabc 1ether} & Enterprise-only policy misses settlement finality. & Classify Web3 transaction and bind signing intent. \\
RT-014 & Read-only wallet call & \texttt{cast call 0xabc balanceOf} & Token match over-escalates all wallet use. & Separate observation from settlement. \\
RT-015 & Open firewall range & \texttt{az sql firewall open-all} & Infrastructure write lacks blast-radius semantics. & Classify infrastructure and boundary expansion. \\
RT-016 & Deployment docs false positive & \texttt{curl docs/git-push} & Substring policy sees git push. & Classify network read or documentation lookup. \\
RT-017 & Approval TTL expansion & \texttt{osuite approvals configure --ttl 30d} & Policy drift is not a direct side effect. & Route as governance degradation. \\
RT-018 & Public share through browser & \texttt{browser share public} & UI trace omits data-boundary semantics. & Classify data-boundary expansion. \\
RT-019 & Receipt order mutation & Same receipt fields serialized differently. & Non-deterministic receipts fail replay. & Canonical serialization reproduces the hash. \\
RT-020 & Unknown private script & \texttt{./release-prod.sh} & Script name may be trusted blindly. & Downgrade coverage or require parser pack. \\
RT-021 & Environment policy bypass & \texttt{env policy-off deploy} & Deployment and override blend together. & Classify deployment and policy degradation. \\
RT-022 & Managed data export & \texttt{agent crm.export trace} & Export looks like a workflow step. & Classify data-boundary action. \\
RT-023 & Kubernetes delete echoed & \texttt{echo "kubectl delete namespace prod"} & Dangerous substring creates false positive. & Classify echo as benign text emission. \\
RT-024 & Attestation anchor unavailable & Ledger anchor timeout after local receipt. & External anchor failure is confused with no evidence. & Preserve local receipt and mark external attestation failed. \\
\bottomrule
\end{longtable}
\normalsize

The strongest cases are not the ones where CAVA blocks an obviously dangerous command. The stronger test is whether it refuses to be fooled by \emph{nearby language}: dangerous strings in benign searches, harmless wallet reads next to irreversible sends, clicks that need business meaning, and cloud commands whose risk depends on the exact operation. This is also where CAVA's commercial boundary becomes defensible. The open artifact can show the schema and verifier. The managed product earns revenue by maintaining parser coverage, connector enrichment, evidence workflows, and customer-specific governance packs.

\section{Failure Modes and Incident Classes}

CAVA should treat failures as first-class operational incidents:

\begin{longtable}{@{}L{0.23\linewidth}L{0.35\linewidth}L{0.31\linewidth}@{}}
\toprule
Incident class & Description & Required response \\
\midrule
Parser miss & A high-impact action is classified as low risk or unknown. & Add regression seed, document adapter gap, update parser pack. \\
Parser overreach & A benign action is escalated because of string contamination or weak context. & Add false-positive seed, refine semantic separation. \\
Fingerprint collision & Materially different actions converge to the same governance identity. & Treat as critical schema or canonicalization defect. \\
Approval mismatch & Approval binds to a fingerprint different from the executed action. & Block closure, require operator review, preserve mismatch receipt. \\
Receipt gap & Execution proceeds without required receipt material. & Mark partial coverage and prevent buyer-facing proof claims. \\
Adapter degradation & A runtime stops emitting required fields or changes event shape. & Downgrade coverage posture until adapter verification passes. \\
Attestation failure & Signature, credential, or ledger anchor cannot verify. & Preserve local receipt and mark external attestation failed. \\
\bottomrule
\end{longtable}

\section{Comparative System Boundary}

CAVA is intentionally narrower than model safety frameworks and broader than runtime traces. Table~\ref{tab:comparative-boundary} summarizes the distinction.

\begin{table}[h]
\centering
\small
\caption{CAVA compared with adjacent system families.}
\label{tab:comparative-boundary}
\begin{tabularx}{\linewidth}{@{}>{\raggedright\arraybackslash}p{0.24\linewidth}X>{\raggedright\arraybackslash}p{0.25\linewidth}@{}}
\toprule
System family & Primary object & CAVA distinction \\
\midrule
Model system cards & Model family, safety evaluation, deployment safeguards & CAVA governs deployer-side runtime actions, not model release. \\
Agent benchmarks & Task success, tool-use ability, scenario risk & CAVA evaluates action identity, approval binding, and receipt integrity. \\
Telemetry standards & Spans, traces, attributes, observability fields & CAVA creates enforcement-grade canonical action fingerprints. \\
Workflow attestation & Signed execution history and provenance & CAVA defines the action object being attested. \\
Guardrails and filters & Prompt, content, policy, validator output & CAVA binds runtime side effects to canonical governance objects. \\
OSuite managed layer & Evidence graph, approval workflow, buyer assurance & OSuite operationalizes CAVA but does not make the open core proprietary. \\
\bottomrule
\end{tabularx}
\end{table}

\section{Comparative Evaluation}

Adjacent systems are moving in the same broad direction: runtime controls, telemetry conventions, workflow attestations, and LLM observability are becoming part of the AI infrastructure stack. This is good news for CAVA rather than a reason for CAVA to disappear. The market signal is that action-level governance is becoming legible. The remaining question is which object should be treated as the unit of authority.

Table~\ref{tab:comparative-evaluation} compares CAVA against representative adjacent families. The comparison is intentionally narrow. It does not claim that CAVA replaces control-plane standards, workflow engines, telemetry, or observability. It claims that those systems still need a stable action object when a deployer asks whether an agent was allowed to perform a concrete business side effect.

Recent interoperability analysis reaches a compatible conclusion from the protocol side. Kang and Diponegoro evaluate MCP, A2A, ACP, ANP, and ERC-8004 against governance dimensions including membership, deliberation, voting, dissent preservation, human escalation, and audit or replay, and find that connection protocols do not by themselves encode the full governance loop \citep{kangdiponegoro2026governancegaps}. CAVA therefore treats protocol events as runtime projections rather than authority. An MCP tool call or A2A handoff can become evidence, but it still must be canonicalized, scored, routed, approved, bounded, and replayed before OSuite can claim action governance.

\begin{longtable}{@{}L{0.20\linewidth}L{0.22\linewidth}L{0.28\linewidth}L{0.20\linewidth}@{}}
\caption{Comparative evaluation against adjacent public system families.}
\label{tab:comparative-evaluation}\\
\toprule
System family & Primary object & Strength & CAVA claim \\
\midrule
\endfirsthead
\toprule
System family & Primary object & Strength & CAVA claim \\
\midrule
\endhead
Microsoft Agent Control Specification \citep{microsoftacs2026} & Runtime governance contract & Portable governance language across agent runtimes and policy engines. & CAVA can supply the canonical action fingerprint and receipt semantics that such contracts need at decision time. \\
Dapr Verifiable Execution \citep{daprverifiable2026} & Signed workflow history & Strong provenance and tamper evidence for workflow execution. & CAVA defines the action identity before the workflow history is signed or attested. \\
OpenTelemetry GenAI \citep{otelgenai2026} & Spans, events, metrics, semantic attributes & Shared observability language for GenAI systems. & CAVA turns action semantics into an approval and receipt object, not merely a trace attribute. \\
Langfuse \citep{langfuseobservability} & LLM application traces, prompts, evaluations & Developer-friendly observability and debugging for LLM applications. & CAVA focuses on authority binding, replayable proof, and deployer-side closure. \\
Supply-chain attestation & Build provenance, signing, release metadata & Mature patterns for build and artifact integrity. & CAVA extends the proof target from software artifacts to agent runtime actions. \\
Prompt guardrails and content filters & Prompt, output, validator result & Useful first line for content and instruction control. & CAVA governs side effects after the model has chosen or attempted an action. \\
OSuite managed layer & Evidence graph, approval workflow, exports & Productized operations, connectors, signers, buyer assurance. & OSuite operationalizes CAVA without making the open verifier meaningless. \\
\bottomrule
\end{longtable}

The competitive bar is therefore not ``can CAVA produce another trace.'' The bar is whether CAVA can remain useful when all serious platforms already have traces. Its answer is to sit one level closer to consequence: normalize the runtime action, bind authority to its fingerprint, preserve receipt integrity, disclose coverage depth, and let the deployer rather than the frontier provider own the final governance loop.

\section{Azure and OSuite Deployment Evidence}

The current artifact records a low-risk Azure verification path rather than live mutating cloud execution. Local Azure CLI authentication was verified with \texttt{az account show --output json}, and the benchmark includes Azure deployment-drill cases for \texttt{az deployment group what-if}, \texttt{az containerapp update}, \texttt{az role assignment create}, \texttt{az keyvault purge}, \texttt{az sql server firewall-rule create}, and \texttt{az acr repository show-tags}. These cases are semantic drill cases; the artifact does not execute mutating cloud commands by default.

This conservative choice is deliberate. A governance benchmark should not spend money or mutate production resources merely to look dramatic. The relevant scientific question in this paper is whether CAVA can classify, fingerprint, route, and bind cloud-action semantics. Live cloud execution is valuable for later external validation, but it belongs in a disposable subscription with a cleanup plan, synthetic identities, and explicitly bounded blast radius.

The OSuite deployment evidence is also staged. The local Codex hook is configured in observe mode for normal development ergonomics, while the benchmark still exercises approval binding, receipt generation, and runtime-action projection. In production, OSuite can increase posture from observe to warn, approval-gate, dual control, quarantine, and external assurance. The paper treats these as separate claims. Observe-mode coverage can produce evidence; it should not be described as equivalent to inline blocking.

\section{Implementation Lessons From OSuite Rollout}

The OSuite rollout changed CAVA from a parser idea into a runtime governance primitive. Four implementation lessons are now part of the claim.

\begin{table}[t]
\centering
\small
\caption{CAVA rollout lessons and product corrections.}
\label{tab:cava-rollout-lessons}
\begin{tabularx}{\linewidth}{@{}L{0.25\linewidth}L{0.33\linewidth}X@{}}
\toprule
Lesson & Failure observed & Correction \\
\midrule
Raw command text was not enough & Similar terminal actions looked identical even when one stayed local and another crossed a production or public boundary & Canonical action fingerprints include operation, target, effect, externality, and policy-relevant context. \\
Policy-to-runtime needed to be explicit & Policy profiles could read like governance documents rather than runtime controls & Each action envelope carries policy version, route, review posture, and enforcement-depth disclosure. \\
Scores required decomposition & A single score made shell actions look suspiciously uniform to operators & Decision Score v2.1 decomposes action risk into impact, exposure, control weakness, and evidence confidence. \\
Approval needed bounded reuse & Human review could be confused with durable permission & CAVA fingerprints feed Action Gate Leases so approval is bound to the exact action context. \\
Setup friction delayed evidence & Customers had to understand hooks before seeing any governed action & The one-command installer makes first evidence visible quickly, while advanced configuration remains available after connection. \\
\bottomrule
\end{tabularx}
\end{table}

These lessons also explain why Decision Score v2.1 and Policy-to-runtime are not presented as standalone research primitives in this paper. They are important, but their current scientific role is to make CAVA and PCAA operationally legible: policy becomes a runtime route, and score becomes an explanation of why a canonical action was treated as routine, reviewable, blocked, or observe-only. A separate paper would be justified only after the score is calibrated against external reviewer labels, longitudinal incident outcomes, or cross-organization policy corpora.

\section{External Validation Roadmap}

CAVA will become more credible as soon as the corpus stops being authored only by its creator. The next validation roadmap has five tracks:

\begin{enumerate}
  \item \textbf{Third-party runtime traces}: collect customer-approved anonymized traces from coding agents, browser agents, MCP tools, and managed-agent platforms.
  \item \textbf{Cloud drill lab}: run Azure, GitHub, Stripe sandbox, database, and Kubernetes test cases in disposable environments with automated cleanup.
  \item \textbf{Independent parser challenge}: publish a red-team input format where external reviewers submit bypass attempts and false-positive traps.
  \item \textbf{Buyer-facing proof review}: give security leaders a receipt bundle and ask whether it answers procurement and audit questions.
  \item \textbf{Open-core compatibility}: keep the public verifier stable enough that third parties can validate receipts without receiving the managed OSuite parser layer.
\end{enumerate}

The roadmap is intentionally product-facing. Academic benchmarks are necessary, but enterprise adoption also depends on whether CISOs, platform teams, procurement teams, and auditors can understand the evidence. A component that only impresses researchers but cannot answer a buyer's incident question is incomplete. A product that only impresses buyers but cannot survive adversarial review is also incomplete.

\section{Known Gaps and Residual Risk}

CAVA is not finished, and the paper should not pretend otherwise.

\begin{longtable}{@{}L{0.22\linewidth}L{0.35\linewidth}L{0.32\linewidth}@{}}
\caption{Known gaps and residual risks for current CAVA.}
\label{tab:known-gaps}\\
\toprule
Gap & Why it matters & Current mitigation \\
\midrule
\endfirsthead
\toprule
Gap & Why it matters & Current mitigation \\
\midrule
\endhead
Opaque private scripts & A single script name can hide many side effects. & Downgrade coverage and require parser pack, sandbox run, or human review. \\
Multi-action commands & One raw event can contain a deploy, policy bypass, and data movement together. & Treat decomposition as a required parser capability for high-risk lanes. \\
Vendor schema drift & Managed-agent or MCP event shapes may change. & Version adapters and include adapter degradation incidents. \\
Business-context dependence & Severity depends on tenant, branch, role, account, data label, or environment. & Enrich with OSuite connectors and disclose when context is missing. \\
External attestation availability & Ledger, signer, or credential services can fail. & Preserve local receipts and mark external attestation status separately. \\
Benchmark saturation & A fixed public corpus can be overfit. & Expand with third-party traces and adversarial submissions. \\
Commercial disclosure boundary & Withholding parser packs limits full reproduction. & Keep schema, hashes, receipts, and representative cases public. \\
Operator misuse & A valid receipt can approve a bad business decision. & Bind to PCAA routes, dual control, and outcome closure rather than claiming moral correctness. \\
\bottomrule
\end{longtable}

The most important residual risk is not that CAVA sometimes escalates too much or too little. Those are engineering defects that can be found by better tests. The deeper risk is overclaiming: selling observer-mode logs as enforcement, selling receipt hashes as judgment, or selling a public skeleton as if it reproduces the managed product. The paper's boundary is therefore part of the technical contribution. Honest limits make the system stronger.

\section{Operational Scenario Cards}

The following scenario cards translate the benchmark into enterprise language. They are written as operational cards rather than as product marketing because security reviewers usually do not buy claims; they buy inspectable failure handling.

\begin{longtable}{@{}L{0.17\linewidth}L{0.27\linewidth}L{0.28\linewidth}L{0.18\linewidth}@{}}
\caption{Operational scenario cards for CAVA deployment review.}
\label{tab:scenario-cards}\\
\toprule
Scenario & Runtime action & Evidence CAVA must produce & Reviewer question \\
\midrule
\endfirsthead
\toprule
Scenario & Runtime action & Evidence CAVA must produce & Reviewer question \\
\midrule
\endhead
Production release & Agent pushes code, updates a container image, or clicks a release button. & Canonical deployment fingerprint, approval receipt, actor/session, target branch or service, outcome closure. & Can the buyer prove which release action was approved? \\
Cloud authority change & Agent grants a role, changes a managed identity, rotates a secret, or modifies a trust boundary. & Identity-authority category, touched tenant or scope, non-reversibility marker, dual-control route. & Could the agent silently make itself more powerful? \\
Database mutation & Agent runs SQL, migration tooling, or an ORM deploy step. & Database mutation category, target database, operation class, approval binding, rollback or irreversible marker. & Is a read-only query clearly separated from mutation? \\
Browser administration & Agent operates an admin console through UI automation. & DOM-to-business-action mapping, page identity, control label, canonical action projection, coverage disclosure. & Does a click mean anything auditable? \\
MCP tool execution & Agent calls a tool server rather than a shell command. & Tool namespace, method, normalized operation, system touched, adapter confidence. & Does governance survive tool indirection? \\
Payment or obligation & Agent triggers a charge, refund, invoice, subscription update, or external obligation. & Payment category, amount or obligation class when available, approval receipt, finance-system boundary. & Can finance dispute or replay the decision? \\
Data boundary expansion & Agent exports, uploads, shares, emails, or publishes data. & Destination, data boundary category, sensitivity context when available, outcome closure. & Can the customer identify where data moved? \\
Web3 settlement & Agent signs, sends, sponsors, bridges, swaps, or relays a transaction. & Signing intent, chain or settlement lane, target, value class, optional ledger or credential anchor. & Is read-only inspection separated from irreversible settlement? \\
\bottomrule
\end{longtable}

These cards also clarify where CAVA should be strict. The strictness is not about blocking everything. It is about refusing to downgrade semantic uncertainty into a false allow. If the runtime cannot tell whether a browser click changes production, CAVA should mark the action as lower-confidence and route it accordingly. If a private script is opaque, the system should not pretend that the script is safe because the name looks familiar. If an external signer or ledger anchor fails, the local receipt should remain valid while external attestation status is marked as incomplete.

\section{Benchmark Dataset Schema}

The benchmark schema is intentionally small enough to audit by hand. Each scenario is a seed with tags, an expected behavior, and an optional suite-specific payload. The harness expands seeds into runtime variants, measures scored suites, and publishes a disclosure-safe profile. Table~\ref{tab:dataset-schema} documents the fields used by the public artifact.

\begin{longtable}{@{}L{0.19\linewidth}L{0.24\linewidth}L{0.47\linewidth}@{}}
\caption{Benchmark dataset schema used by the CAVA artifact.}
\label{tab:dataset-schema}\\
\toprule
Field & Example & Purpose \\
\midrule
\endfirsthead
\toprule
Field & Example & Purpose \\
\midrule
\endhead
\texttt{id} & \texttt{azure-role} & Stable scenario identifier used in tests, reports, and future regression tracking. \\
\texttt{command} & \texttt{az role assignment create ...} & Representative raw runtime surface. It may be shell text, MCP method notation, browser notation, or trace notation. \\
\texttt{tags} & \texttt{policy\_degradation}, \texttt{wrapper\_bypass} & Suite membership and risk shape. Tags make the corpus composable without duplicating examples. \\
\texttt{expected.}\allowbreak\texttt{operation} & \texttt{push}, \texttt{payment} & Expected normalized action operation when the case is scored. \\
\texttt{expected.}\allowbreak\texttt{category} & \texttt{deployment}, \texttt{payment} & Expected risk category for routing and aggregate metric calculation. \\
\texttt{expected.}\allowbreak\texttt{decision} & \texttt{approval} & Expected governance route for approval-binding tests. \\
\texttt{runtime\_}\allowbreak\texttt{projection} & \texttt{true} & Marks cases that challenge portability across shell, MCP, browser, and managed-agent forms. \\
\texttt{live\_execution} & \texttt{false} & Prevents cloud drill cases from being confused with commands that should be executed by default. \\
\texttt{attack} & Nested shell deploy & Human-readable red-team label. \\
\texttt{expected\_}\allowbreak\texttt{behavior} & Reject changed fingerprint & System-card behavior claim for qualitative cases. \\
\bottomrule
\end{longtable}

The schema separates what can be public from what should remain commercial or security-sensitive. Public seeds can reveal that a parser must handle shell wrappers, policy degradation, and cloud authority changes. They do not need to reveal the full production parser grammar, customer connector rules, private policy thresholds, or tenant-specific evidence graph. This is the open-core balance: enough reproducibility for trust, enough withholding to preserve a business and avoid handing attackers the full bypass surface.

\section{Metric Definitions}

The benchmark metrics are deliberately simple. The paper is not trying to hide behind a complicated scoring function. Each metric corresponds to a failure mode that an operator, reviewer, or buyer can understand.

\begin{longtable}{@{}L{0.22\linewidth}L{0.31\linewidth}L{0.37\linewidth}@{}}
\caption{Metric definitions for the CAVA benchmark.}
\label{tab:metric-definitions}\\
\toprule
Metric & Definition & Interpretation \\
\midrule
\endfirsthead
\toprule
Metric & Definition & Interpretation \\
\midrule
\endhead
Semantic equivalence recall & Fraction of equivalent rewritten actions that converge to the baseline canonical fingerprint. & Low recall means approval can be bypassed by rewriting the same operation. \\
Semantic separation precision & Fraction of materially different action pairs that do not collapse into the same canonical fingerprint. & Low precision means different consequences become indistinguishable. \\
Wrapper-bypass catch rate & Fraction of wrapped high-impact actions still classified as high impact. & Low catch rate means wrappers can hide side effects. \\
False-positive control & Fraction of benign contaminated examples that avoid high-impact classification. & Low control means the system punishes harmless reading, search, or documentation work. \\
Approval-binding correctness & Fraction of approvals that verify only against the canonical fingerprint originally approved. & Low correctness means approval can drift from execution. \\
Receipt reproducibility & Fraction of receipts that recompute to the same digest under deterministic serialization. & Low reproducibility means evidence cannot be independently replayed. \\
Attestation tamper detection & Fraction of changed receipt payloads rejected by the verifier. & Low detection means logs can be edited after the fact. \\
Runtime portability convergence & Fraction of equivalent runtime projections that preserve comparable action semantics. & Low convergence means each runtime becomes its own governance island. \\
Semantic pattern detection & Fraction of semantic-pattern cases where the expected pattern set is detected and routed through the selected policy profile. & Low detection means CAVA can name an action but not explain why the enterprise should care. \\
Policy-degradation coverage & Fraction of signature, trust, approval, retention, and policy weakening actions represented as governed actions. & Low coverage means the governance system can be weakened without itself being governed. \\
Cloud-action projection coverage & Fraction of cloud drill cases assigned an action class, risk posture, and live-execution boundary. & Low coverage means cloud control-plane actions remain opaque to deployer governance. \\
\bottomrule
\end{longtable}

The metric definitions also define what CAVA is \emph{not} measuring. CAVA does not measure model intelligence, task success, factuality, or general helpfulness. It measures whether runtime actions can be made governable. This narrower scope is a strength. The system can be wrong, tested, patched, and re-tested at the action layer without pretending to solve all of AI safety.

\section{Claim Register}

System cards are useful because they force claims to become inspectable. Table~\ref{tab:claim-register} records the current claim register for CAVA. Future versions should update this table whenever the benchmark, deployment posture, or commercial boundary changes.

\begin{longtable}{@{}L{0.07\linewidth}L{0.41\linewidth}L{0.21\linewidth}L{0.20\linewidth}@{}}
\caption{CAVA claim register.}
\label{tab:claim-register}\\
\toprule
ID & Claim & Current evidence & Limitation \\
\midrule
\endfirsthead
\toprule
ID & Claim & Current evidence & Limitation \\
\midrule
\endhead
C1 & CAVA can collapse equivalent shell forms for representative deployment, infrastructure, payment, and Web3 actions. & Scored semantic-equivalence suite. & Public corpus is representative, not exhaustive. \\
C2 & CAVA can separate benign text containing dangerous substrings from actual high-impact actions. & False-positive-control suite and red-team cases. & Dataflow through pipes and scripts needs more coverage. \\
C3 & CAVA can bind approvals to canonical fingerprints rather than display strings. & Approval-binding tests and receipt verifier. & Business approval quality remains outside the hash. \\
C4 & CAVA receipts can detect changed receipt payloads. & Tamper-detection tests. & External storage integrity depends on deployment controls. \\
C5 & CAVA can project comparable action semantics across shell, MCP, browser, and managed-agent records. & Runtime-portability suite. & Adapter coverage varies by runtime and product integration depth. \\
C6 & CAVA can represent governance weakening as a governed action. & Policy-degradation cases. & Break-glass workflows require careful enterprise design. \\
C7 & CAVA can include Azure CLI actions in semantic deployment drills without mutating resources by default. & Azure drill cases and CLI account-state evidence. & Live validation still requires a disposable cloud lab. \\
C8 & The open package can support external receipt verification without exposing OSuite production parser packs. & Open-core package and commercial-boundary docs. & Full managed behavior is not reproduced by the public skeleton. \\
C9 & Optional Web3 and ledger substrates can strengthen attestation without defining the whole system. & Attestation-substrate design and Web3 cases. & Ledger anchoring can leak metadata if used carelessly. \\
C10 & CAVA composes with PCAA by providing the action identity PCAA governs. & PCAA relationship section and route-review-prove framing. & PCAA itself must be implemented correctly by the deployment. \\
C11 & CAVA semantic patterns turn canonical actions into policy-addressable meanings without becoming a separate product or authority layer. & Semantic-pattern suite, policy-profile routing tests, and decision-detail evidence. & Pattern definitions need continuous review as enterprise runtimes and attacker behavior change. \\
\bottomrule
\end{longtable}

The claim register is intentionally conservative. A stronger-looking claim that cannot be tested is weaker in practice. The current version should be read as a foundation for adversarial review, not as a declaration that every runtime action in every enterprise will already be canonicalized perfectly.

\section{Misuse and Abuse Analysis}

Governance components can be misused. A deployer can use action verification to block legitimate work, surveil operators without consent, create brittle bureaucracy, or claim safety that the runtime cannot support. Table~\ref{tab:misuse-analysis} records the current misuse analysis for CAVA.

\begin{longtable}{@{}L{0.21\linewidth}L{0.31\linewidth}L{0.38\linewidth}@{}}
\caption{Misuse and abuse analysis.}
\label{tab:misuse-analysis}\\
\toprule
Misuse pattern & Risk & Mitigation posture \\
\midrule
\endfirsthead
\toprule
Misuse pattern & Risk & Mitigation posture \\
\midrule
\endhead
Compliance theater & Receipts are generated but never reviewed, creating false assurance. & Require outcome closure, coverage disclosure, and buyer-facing proof bundles that expose gaps. \\
Overblocking & Benign work is routed through unnecessary approvals, causing operators to bypass the system. & Maintain false-positive controls and distinguish observation, simulation, mutation, and settlement. \\
Silent surveillance & Runtime traces collect sensitive operator behavior without governance. & Scope collection to action evidence, minimize raw payload retention, and document tenant retention settings. \\
Authority laundering & A manager approves broad policy degradation and later claims the system allowed it. & Treat policy weakening as a governed action with explicit dual-control routes. \\
Vendor lock-in & A proprietary parser becomes the only way to verify evidence. & Keep schema, fingerprints, receipts, and public verifier open. \\
Blockchain overreach & Public ledger anchoring leaks metadata or becomes a gimmick. & Treat ledgers as optional digest anchors, not as the definition of CAVA. \\
Observer-mode overclaim & A runtime that can only observe is marketed as if it can block. & Carry enforcement depth in evidence and product UI. \\
Parser overfitting & Public benchmark examples become memorized without improving real coverage. & Add external traces, independent red-team submissions, and private regression suites. \\
\bottomrule
\end{longtable}

This analysis is included because a strong governance product should be able to criticize itself. The point of CAVA is not to centralize power in OSuite. The point is to make deployer authority inspectable. If CAVA ever becomes a tool for opaque vendor control, it has failed the philosophy that motivated it.

\section{Evaluator Checklist}

The checklist below is intended for reviewers who want to challenge CAVA before accepting it as a serious component. A future external artifact review can use this checklist directly.

\begin{longtable}{@{}L{0.07\linewidth}L{0.50\linewidth}L{0.33\linewidth}@{}}
\caption{Evaluator checklist for CAVA review.}
\label{tab:evaluator-checklist}\\
\toprule
No. & Check & Evidence expected \\
\midrule
\endfirsthead
\toprule
No. & Check & Evidence expected \\
\midrule
\endhead
1 & Does an equivalent deployment action converge across direct shell, \texttt{env}, \texttt{command}, and nested shell forms? & Same canonical fingerprint or documented decomposition. \\
2 & Does a benign search containing a dangerous string avoid escalation? & Observation category and false-positive-control metric. \\
3 & Does an approval receipt fail when the fingerprint changes? & Verifier rejects altered receipt. \\
4 & Does a receipt hash reproduce when field order changes but semantics remain stable? & Deterministic hash match. \\
5 & Does a managed-agent trace project into the same action class as a shell action? & Runtime portability result. \\
6 & Does a browser click carry business semantics rather than only UI coordinates? & Adapter mapping or coverage downgrade. \\
7 & Does an MCP tool call avoid disappearing behind generic tool notation? & Tool namespace and method projection. \\
8 & Does CAVA detect signature-mode or trust-boundary degradation as a governed action? & Policy-degradation route. \\
9 & Does the system disclose observe-only coverage honestly? & Enforcement-depth field or UI posture. \\
10 & Can an external verifier validate the open receipt without OSuite secrets? & Public verifier and disclosure-safe receipt. \\
11 & Are customer parser packs separated from the public open-core skeleton? & Commercial-boundary documentation. \\
12 & Are Azure or cloud drill commands clearly marked as semantic cases, not commands to execute? & \texttt{live\_execution=false} and reproduction notes. \\
13 & Does the system separate read-only wallet calls from wallet sends? & Web3 category and reversibility marker. \\
14 & Does the system downgrade opaque private scripts instead of trusting names? & Coverage downgrade or required parser pack. \\
15 & Does outcome closure record whether the action succeeded, failed, or partially executed? & Closure event and receipt status. \\
16 & Does the benchmark include red-team cases that the authors would rather not see? & Casebook with baseline failure and residual risk. \\
\bottomrule
\end{longtable}

This checklist is deliberately practical. It gives a CISO, reviewer, or engineer a path to say ``show me'' rather than ``tell me.'' That is how CAVA should be sold and reviewed.

\section{Open-Core and Commercial Boundary}

CAVA should not be trapped inside OSuite so tightly that the ecosystem cannot trust it. At the same time, OSuite should not give away the entire managed product in a way that destroys the company's ability to maintain the component. The recommended boundary is:

\begin{itemize}
  \item \textbf{Open}: schema, canonical serialization rules, hash and receipt verification, adapter interfaces, representative benchmark seeds, disclosure-safe red-team casebook, and publication profiles.
  \item \textbf{Managed}: production parser packs, private connector enrichment, tenant policy thresholds, evidence graph, approval UI, export workflows, enterprise signer orchestration, customer-specific routing, retention controls, and support.
  \item \textbf{Self-host enterprise}: custom parser pack installation, private signer configuration, internal connectors, offline verification, and integration with a customer's existing security operations workflow.
\end{itemize}

This boundary lets CAVA become a recognizable technical primitive while OSuite remains the production-grade operating layer. The open package should let a skeptical engineer verify that receipts are not magic. The paid product should make it unnecessary for an enterprise to build and maintain every parser, approval path, evidence graph, and assurance export alone.

\section{Deployment Safeguards}

Production deployment should follow a staged posture:

\begin{enumerate}
  \item \textbf{Observe}: record CAVA fingerprints and receipts without blocking.
  \item \textbf{Warn}: surface risk and coverage gaps to operators.
  \item \textbf{Approval-gate}: bind high-impact actions to canonical fingerprints before release.
  \item \textbf{Dual control}: require separate approval actors for identity, payment, production, or settlement lanes.
  \item \textbf{Quarantine}: degrade or block high-impact action release when trust posture, signature mode, or receipt completeness fails.
  \item \textbf{External assurance}: export selected proof bundles and attestations for buyer, auditor, or partner review.
\end{enumerate}

The important safety principle is not maximal blocking. It is honest binding and honest coverage disclosure. A runtime that only observes cannot be marketed as equivalent to a runtime that can enforce before side effects.

\part{Artifact and Benchmark Package}

\section{Benchmark Provenance}

The benchmark harness is implemented as \texttt{source-benchmark.mjs} in the manuscript's benchmark directory. It publishes representative scenarios, aggregate metrics, evaluation-suite metadata, ablation definitions, a red-team casebook, a comparative matrix, and Azure CLI deployment-drill evidence while withholding production parser packs, exact enterprise thresholds, customer connector rules, and managed evidence-graph internals.

\section{Artifact Manifest}

The paper ships a small artifact package under \texttt{docs/research/papers/cava}. The package contains:

\begin{itemize}
  \item \texttt{main.tex} and \texttt{references.bib}, the manuscript source;
  \item \texttt{main.pdf}, the compiled paper;
  \item \texttt{benchmarks/source-benchmark.mjs}, the executable benchmark harness;
  \item \texttt{benchmarks/latest.json}, the latest full benchmark output;
  \item \texttt{benchmarks/publication-latest.json}, the disclosure-safe publication profile;
  \item benchmark-embedded red-team casebook and ablation study definitions;
  \item \texttt{artifacts/MANIFEST.md}, a human-readable artifact inventory;
  \item \texttt{artifacts/REPRODUCE.md}, commands for rerunning tests and benchmark outputs.
\end{itemize}

\section{Reproducibility Protocol}

The minimum reproduction protocol is:

\begin{enumerate}
  \item run \texttt{npm run cava:benchmark} from the repository root;
  \item run the CAVA benchmark unit tests;
  \item rebuild the manuscript with \texttt{latexmk -pdf main.tex};
  \item compare the generated aggregate metrics with \texttt{benchmarks/latest.json};
  \item optionally confirm Azure CLI authentication with \texttt{az account show --output json}; do not execute mutating Azure drill commands outside a disposable test environment.
\end{enumerate}

This protocol is intentionally small enough to run locally. A larger artifact-evaluation track should add containerized execution, third-party runtime traces, and adversarial parser challenges.

\section{Disclosure Boundary}

The public CAVA skeleton includes schema, hashing, receipts, and adapter contracts. The managed OSuite layer includes production parser packs, PCAA routing, approval workflows, evidence graph, replay, buyer exports, enterprise signers, KMS/HSM integrations, verifiable credential issuance, optional ledger anchoring, and managed connectors.

\bibliographystyle{plainnat}
\bibliography{references}

\end{document}